# Table Comprehension in Building Codes using Vision Language Models and Domain-Specific Fine-Tuning

Mohammad Aqib[1], Mohd Hamza[2], Ying Hei Chui[1], Qipei Mei[1]

*Department of Civil & Environmental Engineering, University of Alberta, Edmonton, Alberta, Canada[1]*

*Department of Electrical Engineering, Aligarh Muslim University, Aligarh, Uttar Pradesh, India[2]*

**Abstract:** Building codes contain critical information for ensuring safety, regulatory compliance, and informed decision-making in construction and engineering. Automated question answering systems over such codes enable quick and accurate access to specific regulatory clauses, improving efficiency and reducing errors. Retrieval-Augmented Generation (RAG) systems are essential for this task as they combine the precision of information retrieval with the generative capabilities of language models. However, tabular data are challenging to extract as they often involve complex layouts, merged cells, multi-row headers, and embedded semantic relationships that are not easily captured by traditional natural language processing techniques and Vision Language Models (VLMs). This paper explores and compares two methods for extracting information from tabular data in building codes using several pre-trained VLMs. First, a direct input method is used, where the image of the page is input directly into the VLMs, which are then tasked with answering questions based on the image. Second, an indirect input method is introduced, which involves converting an image of a page containing tables into the LaTeX code and then answering inquires based on the LaTeX-based input. The experiments find that the direct input method generally resulted in higher accuracy than the indirect input method. To further improve the performance, we fine-tuned each VLM using Low Rank Adaptation (LoRA) on a domain-specific tabular dataset. The fine-tuned models exhibited substantial improvements, with Qwen2.5-VL-3B-Instruct achieving relative accuracy gains exceeding 100%. Our results highlight the potential of parameter-efficient fine-tuning methods to adapt powerful VLMs for understanding complex structured data in specialized fields, such as building code interpretation and regulatory compliance.

# 1. Introduction

Building codes serve as the cornerstone of modern construction, establishing the minimum standards necessary to ensure the safety, health, and structural integrity of buildings for their occupants [1]. These regulations dictate a wide array of requirements, including design of structures [2], fire hazards [3], conservation of energy [4], and the capacity of buildings to withstand natural disasters such as earthquakes [5], wind, and rain [6]. The evolution of these codes is a continuous process informed by the analysis of building failures during catastrophic events or by extensive review of research [7]. In Canada, for instance, after proper research and exploration, insurance stakeholders analysed and proposed a few changes in Part 9 of the building code that governs the construction of low-rise dwellings and makes buildings more resilient to high winds [8]. The fundamental importance of these regulations underscores the critical need for professionals to have efficient and accurate access to their vast content. However, the sheer volume and intricate nature of building code documents present a significant hurdle for professionals in the architecture, engineering, and construction (AEC) industry [9]. Locating specific information or answering queries related to building regulations often becomes a challenging and time-consuming manual task. This difficulty arises not only from the extensive length of these documents but also from the inherent characteristics of their content. To address these inherent difficulties, the development of automated question-answering (QA) systems offers a promising solution [10]. Such systems can provide efficient and rapid access to critical information contained within building codes, potentially revolutionizing how AEC professionals interact with these essential regulations.

A promising approach to building such automated QA systems is multimodal Retrieval-Augmented Generation (MRAG). Retrieval-Augmented Generation (RAG) represents a sophisticated approach to question answering that combines the strengths of information retrieval and generative language models [11]. This methodology first retrieves relevant information from a vast knowledge base in response to a user's query and then utilizes a generative language model to formulate an accurate and contextually appropriate answer based on the retrieved content. MRAG extends this framework by integrating various data formats, including text, images, and videos [12]. The information retrieval component of an MRAG system plays a crucial role in accessing the specific sections of building code documents that are relevant to a user's query [13].

The generative language model in an MRAG system then takes the retrieved information and the original user query as input to construct a comprehensive answer in natural language. MRAG is widely adopted in several fields, and few works include a comprehensive analysis for diseases such as Alzheimer's [14], industrial applications [15], and medical radiology report generation [16]. An overview of the workflow for MRAG is depicted in Figure 1. Among the components, Vision Language Models (VLMs) represent a significant advancement in artificial intelligence, integrating computer vision and natural language processing by jointly learning representations from both visual and textual data [17]. Recent advancements in deep learning and natural language processing have significantly propelled the field of VLMs, leading researchers to develop diverse architectures and techniques that bridge the gap between visual perception and language understanding, ultimately enabling more effective and user-friendly multimodal systems [18]. A few of the renowned VLMs are Llama by Meta [19], InternVL by OpenGV Lab [20], GPT4 by OpenAI [21] , and Qwen by Alibaba [22].

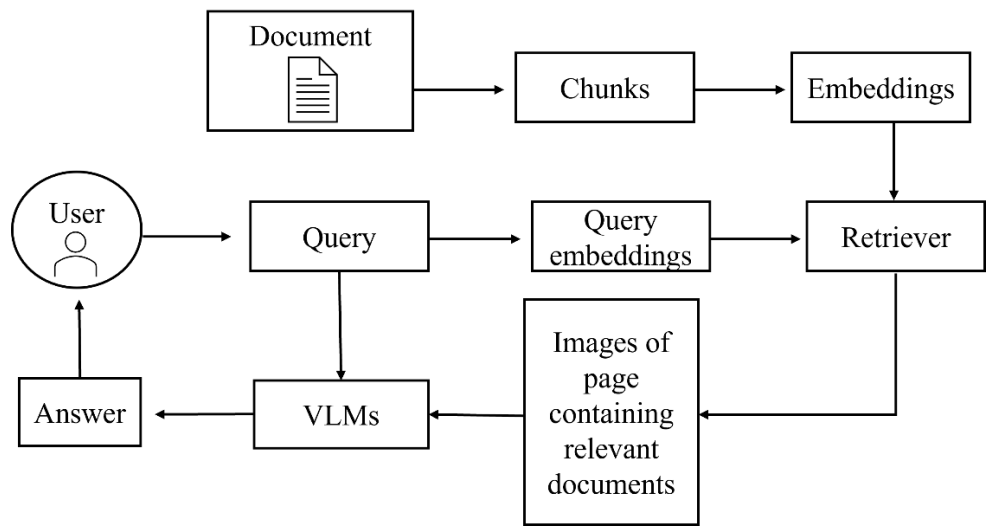

**Figure 1.** Workflow of MRAG.

The core strength of VLMs lies in their ability to process and understand information from diverse data formats, including both images and natural language text [23]. This allows them to perform a wide range of multimodal tasks, such as visual caption restoration [20], Visual Question Answering (VQA) [24], image retrieval based on textual queries [25], and understanding the content present in charts [26]. Their capacity to understand the relationships between visual and textual elements, going beyond mere object recognition to grasp context and semantic connections, makes them particularly powerful for analyzing complex documents [27]. The relevance of VLMs is rapidly increasing across various domains, including civil engineering and construction [28]. In this sector, VLMs are being utilized for tasks such as analyzing architectural drawings to extract design information [29], detecting defects in construction sites through image analysis [30], monitoring construction progress [31], and bridge inspection by processing visual data [32]. The ability of VLMs to understand both visual and textual information makes them exceptionally well-suited for the AEC industry, where information is often conveyed through a combination of drawings, specifications, and regulatory documents.

While VLMs, which are integrated into MRAG, are powerful and continually improving, they may still struggle to achieve high accuracy across a range of tasks. Despite the advancements in VLMs, they still face issues such as object hallucination, text-to-image interference, reduced robustness when confronted with complex concepts, object counting, spatial relation reasoning, and absurd question answering [33]. Specifically, VLMs face challenges when dealing with complex tables [34]. Table detection and recognition is difficult task due to the varying layouts and formats of tables [35]. They can have intricate structures with multi-level headers, spanning cells, and complex formatting [36]. The inability to achieve satisfactory accuracy in interpreting tables can significantly affect the reliability of question-answering systems, especially in cases where critical regulatory information is commonly presented in tabular form, such as building codes.

To address the aforementioned challenges, this paper introduces two approaches, namely direct and indirect input methods, for interpreting images of pages containing tables. The direct input method provides the image directly to the VLMs, which are then tasked with answering questions based on the image itself. In contrast, the indirect input method involves converting such images into LaTeX code, after which inquiries are answered based on the LaTeX-derived content. To further improve the performance of VLMs, this study also employs Low-Rank Adaptation (LoRA), a fine-tuning technique that keeps the original weights of the pre-trained model frozen while introducing a small set of trainable parameters through low-rank matrices [37]. These matrices represent weight updates that enhance the model's ability to accurately identify and extract correct answers from tables.

This paper is organized as follows: Section 2 outlines the methodology, including the evaluation of pre-trained VLMs using different input types, as well as the fine-tuning and testing procedures. Section 3 details the dataset generation process, while Section 4 presents and discusses the results. Finally, Section 5 concludes with the key findings of this study.

## 2. Methodology

### 2.1 Direct and Indirect Input Methods

In this work, we propose processing tabular data in two ways, i.e., direct and indirect input methods. The format in which tabular data is presented to a VLM can significantly influence its ability to comprehend and reason over the information [38]. In the direct input method, images of pages containing tables are sent directly to VLMs to answer queries. This way preserves the table's visual layout and formatting as it appears in the source document. In contrast, In the indirect input method, tables are first converted to LaTeX using a VLM, and the resulting LaTeX is then passed to other VLMs to extract information. LaTeX provides a structured, text-based representation of the table's content and organization. VLMs used for analysing input methods include Llama-3.2-11B-Vision-Instruct, Qwen2-VL-2B-Instruct, Qwen2-VL-7B-Instruct, Qwen2.5-VL-3B-Instruct, and Qwen2.5-VL-8B-Instruct. A detailed overview of the VLMs used is mentioned in Table 1.

**Table 1**. Model used and associated parameters.

| Model Name | Number of parameters (in billion) |
|---|---|
| Llama-3.2-11B-Vision-Instruct | 11 |
| Qwen2-VL-2B-Instruct | 2 |
| Qwen2-VL-7B-Instruct | 7 |
| Qwen2.5-VL-3B-Instruct | 3 |
| Qwen2.5-VL-7B-Instruct | 7 |

The direct input method approach involved applying VLMs directly on the images of pages containing tables without converting them into LaTeX format. For this, we utilized the pre-trained VLMs as presented in Table 1 for answering questions from images of pages containing tables. Figure 2 shows application of direct input method employed in this study.

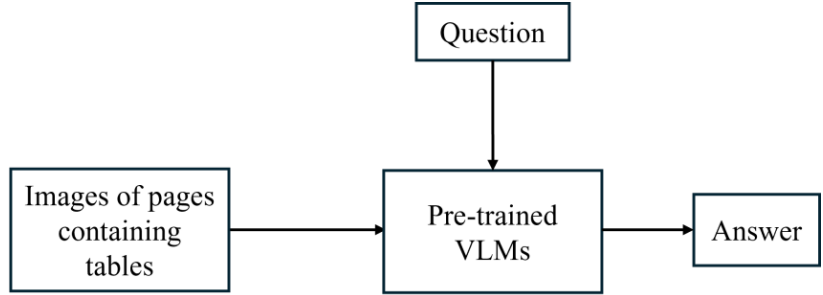

**Figure 2.** Direct input method.

The indirect approach involves applying pre-trained VLMs to the LaTeX representations of tables extracted from building codes. Images of tables were converted into their corresponding LaTeX format using a VLM. This transformation allowed us to leverage the structured format of LaTeX for question-answering task on pre-trained VLMs. GPT-4.1 was chosen to convert images to LaTeX code, as it shows good performance over Massive Multi-discipline Multimodal Understanding and Reasoning (MMMU) Benchmark [39]. An overview of indirect input method is shown in Figure 3.

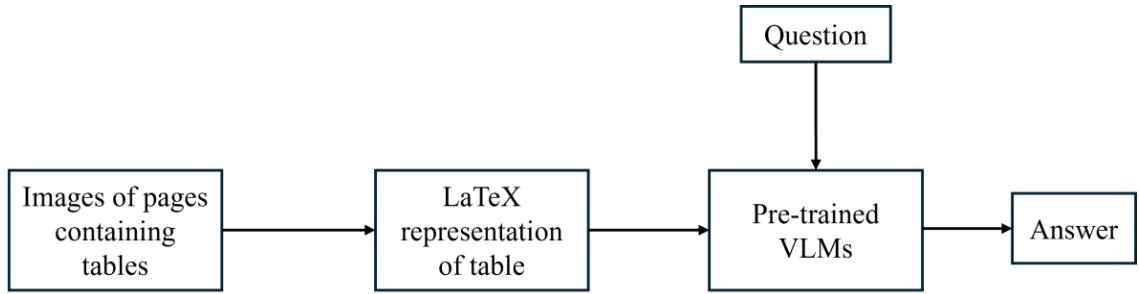

**Figure 3.** Indirect input method.

*2.2 Fine-Tuning VLMs*

VLMs primarily have two key components: a Vision component and a language component. The Vision component encodes images using an image encoder and projects them into a shared embedding space. The language component tokenizes input text, adds positional encodings, and embeds it into the same space. These multimodal embeddings are then passed to a transformer decoder that generates text based on the combined visual and textual context [40]. This enables tasks like image captioning and visual question answering. A basic overview of the architecture of VLM is shown in Figure 4.

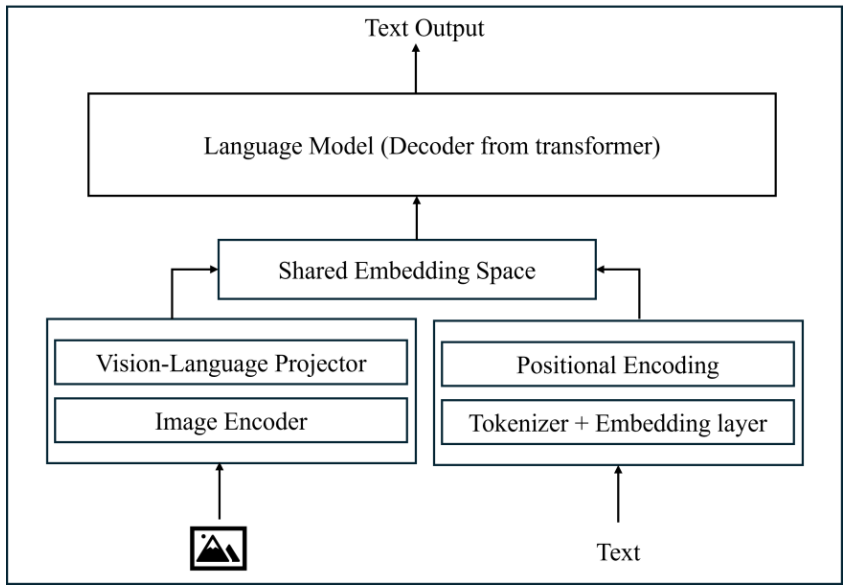

**Figure 2**. Basic architecture of Vision Language Model.

Fine-tuning is a promising solution to improve the capabilities of VLMs to identify and extract the correct answer from table. It involves training pre-trained models on task specific datasets and assists in adapting model on new desired tasks [41]. However, full fledged fine-tuning involves modifying all the parameters of the model and it becomes really difficult when dealing with models with billions of parameters as dependency on computational resources and large dataset increases [42]. However, there exist certain techniques which can align the model with the new task and dataset without the need of heavy computational resources and large datasets. Parameter efficient fine-tuning technique (PEFT) is solution to the challenges faced by full-fledged fine-tuning. Instead of updating all the parameters of the model they freeze most of the parameters and update only few and therefore reduces the constraints which one have to deal with while opting full-fledged fine-tuning [42]. There are several methods of PEFT and they include Visual Prompt Tuning [43], BitFit [44], AdaptFormer [45], Low Rank Adaptation (LoRA) [46]. Another benefit of using PEFT is that even though it modifies only few parameters of model, but still it reports comparable performance to full-fledged fine-tuning [47]. Fine-tuning is a crucial step for adapting a VLM to a particular dataset or to a particular task. In this study, pre-trained VLMs are fine-tuned by utilising direct input method by combining both visual and text inputs present in the prepared QA dataset. Visual input is processed by the vision component that analysed the table's layout and extracted relevant information, while the text was tokenized and passed through the language component of the model. This helps in the model to make connections between the structure of tables and corresponding queries, ultimately assisting in improving the model's ability to answer questions over complex tables. An overview of the fine-tuning process is mentioned in Figure 5.

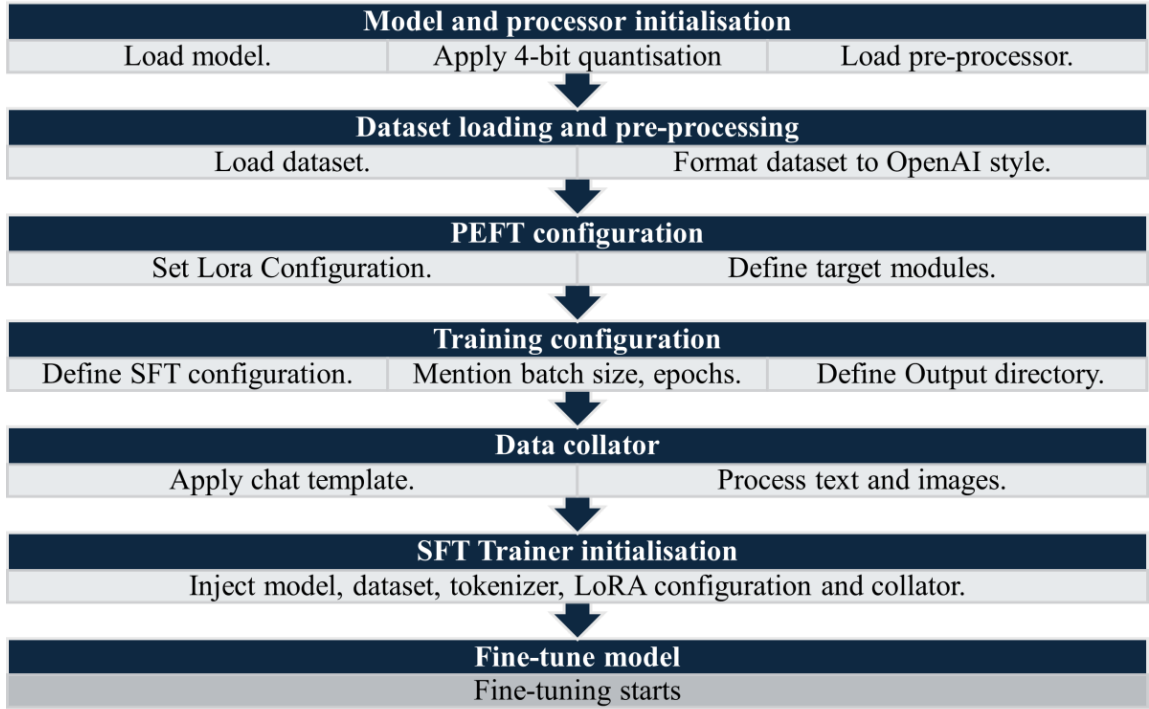

**Figure 3.** Workflow of the fine-tuning process.

Fine-tuning is performed using LoRA, a technique designed to efficiently adapt large pre-trained models to downstream tasks. Instead of updating the full weight matrices, LoRA freezes the original model parameters and introduces a pair of small, trainable low-rank matrices into specific layers. These matrices capture task-specific updates while preserving the knowledge stored in the original weights. This approach significantly reduces the number of trainable parameters, enabling efficient fine-tuning with minimal computational overhead and dataset requirements. The modified weights are computed as:

$$W' = W + \Delta W \qquad (1)$$

$$W' = W + A \cdot B \qquad (2)$$

where A and B are trainable low rank matrices introduced during fine-tuning, W signifies the original weights of the model, ΔW is the change in weights due to fine-tuning, and W' represents updated weights of the model post fine-tuning.

To implement LoRA, BitsandBytesConfig was used, enabling efficient 4-bit quantization of pre-trained VLM. This substantially reduced GPU memory usage while maintaining model performance, making it feasible to fine-tune large-scale VLMs on limited hardware resources. The model was loaded in 4-bit precision, and a corresponding processor was initialized to handle both image and text modalities.

The training dataset extracted from the building codes consists of structured triplets, where each sample includes a table image, a question, and a corresponding ground truth answer. The dataset is formatted into the OpenAI-style message format, which comprises a system prompt, a user message containing both the image and query, and an assistant message with the expected answer. This format ensures seamless integration with the model's chat-style architecture and allowed clearer multi-modal context understanding during fine-tuning. A sample of the training dataset is shown in Figure 6 and a detailed overview of the dataset used is mentioned in Section 3.

**Fasteners for Subflooring and for Sheathing where the 1-in-50 HWP < 0.8 kPa and $S_a(0.2) \leq 0.70$**
Forming Part of Sentence 9.23.3.5.(1)

| Element | Minimum Length of Fasteners, mm | | | | Minimum Number or Maximum Spacing of Fasteners |
|---|---|---|---|---|---|
| | Common or Spiral Nails | Ring Thread Nails or Screws | Roofing Nails | Staples | |
| Board lumber 184 mm or less wide | 51 | 45 | n/a | 51 | 2 per support |
| Board lumber more than 184 mm wide | 51 | 45 | n/a | 51 | 3 per support |
| Fibreboard sheathing up to 13 mm thick | n/a | n/a | 44 | 28 | 150 mm o.c. along edges and 300 mm o.c. along intermediate supports |
| Gypsum sheathing up to 13 mm thick | n/a | n/a | 44 | n/a | |
| Plywood, OSB or waferboard up to 10 mm thick | 51 | 45 | n/a | 38 | |
| Plywood, OSB or waferboard over 10 mm and up to 20 mm thick | 51 | 45 | n/a | 51 | |
| Plywood, OSB or waferboard over 20 mm and up to 25 mm thick | 57 | 51 | n/a | n/a | |

2) Fastening of roof sheathing and sheathing in required *braced wall panels* shall

"Question": "What is the minimum length of fasteners for subflooring and for sheathing required for board lumber 184 mm or less wide when using common or spiral nails?",

"Answer": "51 mm",

"image_file": "table_page_904.png"

**Figure 4.** Snippet of a page displaying a table with its associated question and answer [48].

In the PEFT configuration, we configured LoRA by specifying a targeted subset of transformer components for adaptation. The selected target modules included q_proj, k_proj, v_proj, gate_proj,

down_proj, up_proj, and visual_proj. These modules are central to the model's ability to process both language and vision features, and fine-tuning them enables the model to better learn task-specific interactions between complex table structures and their corresponding textual queries.

- **q_proj, k_proj, v_proj**: These modules are part of the self-attention mechanism and govern how the model attends to different input tokens. Tuning them helps the model better align text descriptions with structured table elements, such as headers, numerical data, and footnotes.

- **gate_proj, down_proj, up_proj**: These are part of the feedforward network within transformer blocks. Modifying these allows for improved abstraction and transformation of modality-specific features.

- **visual_proj**: This module is responsible for projecting visual embeddings into a shared latent space with text embeddings. Fine-tuning this layer ensures that image features, particularly layout, cell structure, and semantic clustering within tables, are accurately represented and aligned with textual context.

This selection of modules is particularly effective for our task, which involves adapting VLMs to handle complex tabular data present in building codes. The dense, rule-heavy structure of tables present in building codes, which are often packed with numeric ranges, merged columns, and multi-row headers, requires the model to extract and correlate multi-modal cues accurately. By selectively adapting the attention, feedforward, and visual projection layers, the model is equipped to build stronger image-text associations and better understand domain-specific tabular layouts.

We employed the SFTTrainer with a training setup consisting of 20 epochs, a batch size of 2, a learning rate of $2 \times 10^{-5}$, and gradient accumulation to simulate a larger effective batch size. Gradient checkpointing and mixed-precision training (fp16) were utilized to manage GPU memory efficiently. An AdamW optimizer was used for parameter updates.

A custom data collator was responsible for batching and structuring the training samples. It applied the model's chat template, tokenized the inputs, and processed the image features using the initialized processor. Padding was applied to standardize input lengths, and both padding and image tokens were masked out in the labels to prevent them from influencing the loss. Finally, the quantized model, tokenizer, processor, LoRA configuration, dataset, and collator were passed into the SFTTrainer for training. Fine-tuning starts with periodic checkpointing to preserve progress. This workflow enabled domain-specific adaptation of VLMs, which are capable of understanding and reasoning over complex regulatory tables in the building codes.

All experiments were carried out on a 64-bit system powered by an 80-core Intel Xeon Gold 6242R CPU clocked at 3.1 GHz, supported by 128 GB of RAM, and running Ubuntu 22.04 (Linux). Local VLMs were inferenced on an NVIDIA A6000 GPU with 48 GB of VRAM. The experimental environment was developed in Python 3.10 and made use of libraries including PyTorch, HuggingFace Transformers, Datasets, PEFT, and TRL.

*2.3 Evaluation*

The overall evaluation will include two phases. In the first phase, the goal was to identify the most effective input method for generating answers by evaluating various pre-trained VLMs using both direct and indirect input methods. For both input methods, a conversational format was used, structured through a message object to simulate a chat-based interaction. This format was adopted to align with the expected input style of VLMs, which are typically trained to understand and respond within dialogue-like contexts. By presenting the queries in a conversational manner, the models could better interpret the user's intent, enabling more coherent and contextually grounded responses. The models were loaded using 4-bit quantization with BitsAndBytesConfig, and torch bfloat16 was utilized as the compute data type to optimize memory usage and accelerate inference. The generated answers from pre-trained models using both input methods were stored and compared with the ground truth using the InternVL3-8B model, as shown in Figure 7. InternVL3-8B evaluates correctness by classifying the answers as either correct or incorrect based on the provided ground truth. InternVL3-8B is a state-of-the-art multi-modal model released by OpenGV Lab, known for its strong reasoning capabilities across both visual and textual modalities. Its effectiveness is further supported by its high performance on challenging benchmarks such as WinoGrande (78.1), HellaSwag (90.2), and BigBench Hard (77.4), which are specifically designed to test commonsense reasoning, situational understanding, and multi-step inference [49]. These benchmark results justify its suitability as an evaluation model in this study. If the generated answer is correct when compared with the ground truth, it is marked as correct; otherwise, if it fails to answer the question correctly and provides a wrong answer, it is marked as incorrect.

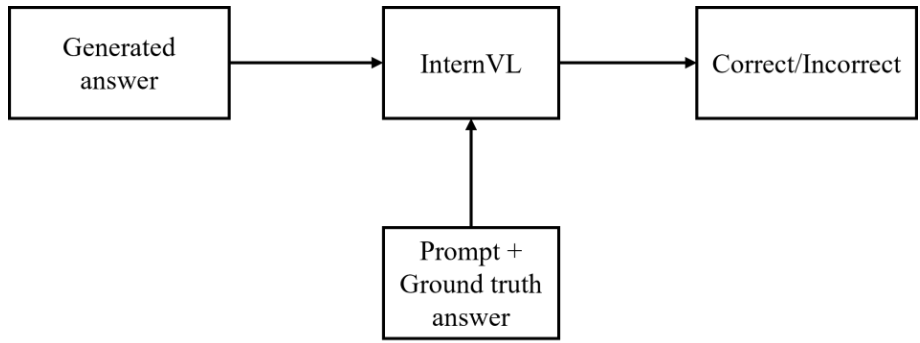

**Figure 5.** Evaluation of the generated answer.

In the second phase of evaluation, the performance of fine-tuned VLMs was tested by generating answers using the direct input method. The fine-tuned models were adapted using LoRA, trained on a domain-specific dataset containing image-question-answer triplets. During inference, the LoRA adapters were applied to the pre-trained models using the load_adapter function, enabling task-specific adaptation without modifying the core model weights. The direct input method was used for generating responses from fine-tuned models, allowing us to evaluate the impact of fine-tuning on model performance. Generated responses from the fine-tuned model were stored in a JSON file for evaluating the correctness of the answer. An overview of structure of JSON for fine-tuned generation is shown in Figure 6. After generating answers, the "InternVL3-8B" model was used to compare the correctness of the answers to the ground truth, as shown in Figure 8.

```
{
    "Question": "What is the maximum building area for a building with 2 storey facing 2 street in Group A, Division 3?",
    "Ground Truth": "2500 m\u00b2\"\n}\n```",
    "Fine-tuned Generation": "2500"
},
{
    "Question": "What is minimum equivalent thickness for monolithic concrete and concrete panels made with Type S concrete at fire resistance rating of 3 hours?",
    "Ground Truth": "158 mm",
    "Fine-tuned Generation": "The minimum equivalent thickness for monolithic concrete and concrete panels made with Type S concrete at a fire-resistance rating of 3 hours is **158 mm**."
},
{
    "Question": "What is the minimum fire-resistance rating for mezzanine floors in a Group C residential building with a maximum building height of 3 storeys?",
    "Ground Truth": "45 min",
    "Fine-tuned Generation": "45 min"
},
```

**Figure 8.** Structure of JSON file with Question, Ground truth and Fine-tuned Generation

To quantify pre-trained model performance, accuracy was computed for different input methods across different models. Accuracy was also used as basis to evaluate the effect of fine-tuning on VLMs. It was calculated using the formula:

$$\text{Accuracy} = \frac{\text{Number of Correct Predictions}}{\text{Total Number of Predictions}} \quad (3)$$

This metric provided a clear measure of how often the model's output matched the expected ground truth. Additionally, confusion matrices were generated to visualize the distribution of predictions, enabling a deeper analysis of performance differences between pre-trained and fine-tuned models. The confusion matrix not only helps quantify the agreement and disagreement between the pre-trained and fine-tuned models but also highlights the specific cases where fine-tuning improved or regressed performance. This evaluation strategy offered insights into the effectiveness of both direct and indirect input methods and the performance gains achieved through domain-specific fine-tuning.

## 3. Dataset

*3.1 Introduction of NBCC*

In this paper, the proposed methods are tested and evaluated on the dataset generated from the National Building Code of Canada (NBCC) 2020. The NBCC 2020, developed by the Canadian Commission on Building and Fire Codes (CCBFC) and published by the National Research Council of Canada (NRC), outlines the technical standards for building design and construction throughout Canada [48]. The CCBFC, an independent advisory body established by the NRC, comprises volunteers from various sectors of the code-using community, including engineers, architects, builders, skilled tradespeople, and fire and building officials. The NBCC offers guidance not only for new construction but also for the renovation, repurposing, and demolition of existing buildings. It plays a key role in ensuring safety, accessibility, and energy efficiency, making it an essential resource for professionals such as architects, engineers, and regulatory personnel. The NBCC is divided into two volumes and structured into three divisions. Division A defines the Code's scope, goals, and functional objectives, offering a conceptual framework rather than direct compliance instructions. Division B presents the technical requirements and acceptable solutions, reflecting the minimum performance standards tied to Division A's objectives, and is the section most commonly used. Division C addresses the administrative rules for applying and enforcing the Code.

*3.2 Dataset curation for direct input method comparison and VLM fine-tuning*

To efficiently extract tabular data from the NBCC, a two-step procedure was followed. In the first step, images of pages containing tables were extracted from the NBCC. In the second step, InternVL2_5-8B was employed to generate Question-Answer (QA) pairs from each extracted image. The choice of InternVL2_5-8B is supported by its strong performance on several benchmarks that test multimodal understanding and reasoning over structured data, such as 93.0 on SEED-2 Plus, 32.9/68.6 on CharXiv (RQ/DQ), 69.7 on ChartQA, and 77.6 on InfoVQA, demonstrating its capability to accurately interpret and reason over document images containing tables and text [20]. Figure 9 shows the steps that were taken to generate a QA pair from the tables present in NBCC. Extraction was handled by a robust Python library named "pdflumber". The algorithm iterated over vast NBCC and extracted images of pages with tables in high resolution. Conversion of images in high resolution was handled by PyMuPDF which sets zoom scaling factor to 3.0. Zoom scaling factor increases the resolution of the image to 300 DPI, which is suitable for both manual validation and OCR. The processed images were later saved in a local directory for further utilisation.

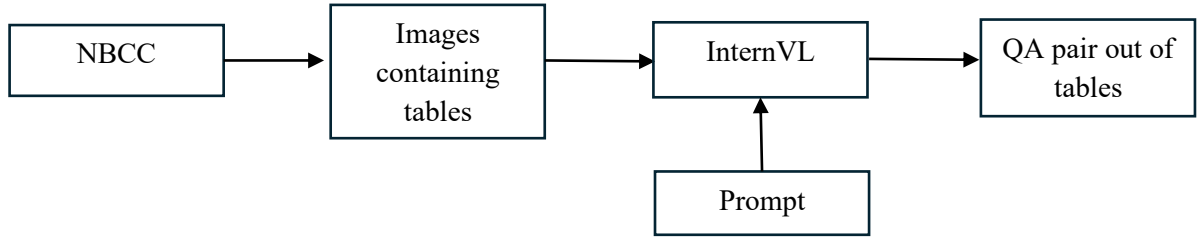

**Figure 9.** Generation of QA pairs out of tables.

Extracted images were used to generate QA pairs using InternVL2_5-8B. Before feeding images to the model, they were passed to a pre-processing pipeline. The pre-processing pipeline converts the images to RGB format and resizes the images while preserving aspect ratios, followed by converting the images into tensors for efficient model input. The model was loaded with BF16 precision to optimise memory usage and inference speed. A carefully crafted prompt was formulated to guide the model in extracting QA pairs from images containing tables. The prompt enforces strict adherence to tabular content, ensuring that the generated QA pairs remain contextually relevant and are generated in well-structured "Question" and "Answer" fields in JSON format. The prompt used for the generation of the dataset is mentioned in Figure 10. Each QA pair was linked to its respective image filename, which facilitates traceability, manual inspection, testing, and fine-tuning VLMs. All the results were aggregated and saved in a structured format. The dataset was further inspected and expanded through manual validation. At the end, we obtained 500 Image-Question-Answer triplets, of which 400 were used for fine-tuning the pre-trained VLMs, and the remaining 100 were reserved for testing the fine-tuned models and analyzing the effectiveness of each input method.

```
prompt = ( """ <image> You are an expert in analyzing complex tabular data. Your task
is to generate highly relevant, precise, and structured Question-Answer (QA) pairs based
solely on the table present in the image. Follow these strict guidelines to ensure
accuracy:

### **1. Focus Exclusively on the Table** - Analyze only the data within the table and
ignore any surrounding text. - Two QA pair out of an image is sufficient.

### **2. Identify Key Relationships in the Table** - Understand how different columns
and rows interact (e.g., material type, grade, supported length, beam size). - Ensure
questions consider dependencies between multiple attributes rather than extracting
isolated values.

### **3. Generate High-Quality Questions** - Formulate questions that accurately
reflect the table's structure and logic. - Ensure that each question is unambiguous and
directly answerable using the provided table.

### **4. Provide a Precise Answer** - Extract the exact value from the table without
additional explanations or justifications. - If multiple values match, return them in a
structured way (e.g., as a list).

### **5. Output Format** - The response must be a **valid JSON object** with two
keys: "Question" and "Answer". - Do not include any additional text, explanations, or
formatting outside this JSON structure. #### **Example Output:**

json

{

"Question": "What is the maximum span for 38×235 beam using Spruce-Pine-Fir (Select
Structural) with a supported length of 3.6 m?",

"Answer": "3.64 m"

}
"""
)
```

**Figure 10.** Prompt used for Image-QA pair generation.

*3.3 Dataset curation for indirect input method*

To convert table images from the NBCC into structured LaTeX code, we employed OpenAI's most advanced vision-language model available at the time of experimentation, GPT-4.1-2025-04-14. This model is capable of processing both image and text inputs in a multi-modal setting. A testing dataset with a total of 100 triplets was used for this task. Images of pages containing tables present in the test dataset were converted to their respective LaTeX format using GPT4.1. Each image was paired with a standardized prompt instructing the model to convert the visual tabular content into LaTeX format. The model returned only the LaTeX code corresponding to the table structure. Finally, a JSON file containing 100 samples of Question-Answer pairs along with the LaTeX representation of the tables was stored for further testing of the indirect input method employed in this study.

## 4. Results and Discussion

*4.1 Analysis of pre-trained VLMs on direct and indirect input methods*

This section details the performance evaluation of five pre-trained VLMs on table understanding tasks, utilizing both direct and indirect input methods. Accuracy percentages for both input types were calculated and are presented in Figure 11.

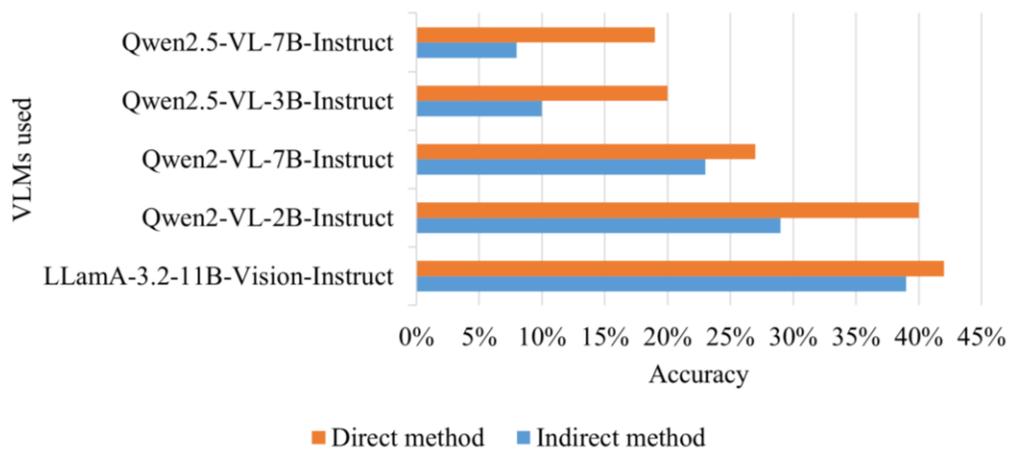

**Figure 116.** Comparison of different models' accuracy on the direct and indirect input methods

The results reveal significant variations in accuracy across the models and methods of input type. For the indirect input method, accuracy ranged from a high of 39% achieved by LLaMA-3.2-11B-Vision-Instruct down to a low of 8% for Qwen2.5-VL-7B-Instruct. A similar pattern was observed for the direct input method, where LLaMA-3.2-11B-Vision-Instruct again achieved the highest accuracy at 42%, while Qwen2.5-VL-7B-Instruct recorded the lowest at 19%. Based on the analysis, Llama-3.2-11B-Vision-Instruct likely outperformed all Qwen models due to its vision adapter with cross-attention, enabling more nuanced interaction between the table image and the question [50]. Additionally, its extensive training on a large multimodal dataset might have included more relevant examples for understanding tables. Finally, its instruction tuning might have been more specifically geared towards visual question answering on structured data like tables.

Another interesting trend emerged within the Qwen model families. Contrary to expectations that newer might perform better, the older Qwen2 series generally outperformed the newer Qwen2.5 series on this specific task. Specifically, Qwen2-VL-2B-Instruct (29% on indirect input method, 38% on direct input method) and Qwen2-VL-7B-Instruct (23% on indirect input method, 27% on direct input method) both achieved higher accuracies than Qwen2.5-VL-3B-Instruct (10% on indirect input method, 20% on direct input method) and Qwen2.5-VL-7B-Instruct (8% on indirect input method, 19% on direct input method). This outcome may be attributed to several factors. First, Qwen2-VL models use Vision Transformer (ViT) derived from Data Filtering Networks (DFN)-based vision encoder, making them better suited for tasks involving complex table layout understanding [51]. In contrast, Qwen2.5-VL employs native dynamic-resolution ViT trained from scratch, with Window Attention for efficiency, and it may lack exposure to structured visual like tables [52]. Additionally, within each Qwen sub-family, smaller model variants demonstrated higher accuracy than their larger counterparts. Qwen2-VL-2B (29% on indirect input method, 40% on direct input method) outperformed Qwen2-VL-7B (23% on indirect input method, 27% on direct input method), and Qwen2.5-VL-3B (10% on indirect input method, 20% on direct input method) outperformed Qwen2.5-VL-7B (8% on indirect input method, 19% on direct input method). This could be due to factors like overfitting in larger models or differences in training dynamics and data exposure.

Consistent superiority of accuracy using the direct input method over accuracy using the indirect input method across all models is a noteworthy finding. This indicates that, for this evaluation, the models were generally more adept at interpreting the images of table and content present in it than understanding the underlying LaTeX code representing the same table. This might imply that the visual processing components and the integration of visual features are relatively more effective. Moreover, it shows that the syntax of LaTeX poses a greater challenge for these model's current language understanding capabilities. Visual cues like cell alignment, borders, and spatial relationships in images might provide more direct information than the corresponding LaTeX commands.

One of the key reasons behind the lower performance on the indirect input method is the inherent complexity and abstraction of LaTeX syntax. Unlike the direct input method, which visually encode table structure, LaTeX represents tables through a series of textual commands that require multi-step parsing, reasoning about alignment, nested formatting, and token ordering. Most pre-trained VLMs may not have been explicitly trained on complex LaTeX table formats, resulting in weaker token-level comprehension of structural semantics. Additionally, GPT-4.1 was employed to convert images of pages containing tables. However, despite being one of the most advanced models, it had struggled in accurately converting the complex tables from the NBCC into their corresponding LaTeX formats. Errors introduced during these conversions, such as misaligned columns, incorrect syntax, or missing formatting, can be another reason behind low performance of pre-trained VLMs on the indirect input method.

Although the direct input method consistently outperformed the indirect input method across all tested models, the overall accuracy remains relatively low. This highlights the limitations of current VLMs in understanding tabular content and underscores the need for domain-specific adaptation. To address this, we fine-tuned pre-trained VLMs on a tabular training dataset derived from the NBCC, aiming to enhance their performance on structured document understanding tasks.

*4.2 Comparing the performance of pre-trained and fine-tuned VLMs*

This section presents a comparative evaluation of pre-trained and fine-tuned VLMs on the test dataset derived from NBCC. The models evaluated include LLaMA-3.2-11B-Vision-Instruct, Qwen2-VL-2B-Instruct, Qwen2-VL-7B-Instruct, Qwen2.5-VL-3B-Instruct, and Qwen2.5-VL-7B-Instruct. Each model underwent fine-tuning using a dataset of 400 domain-specific samples composed of structured question-answer pairs with corresponding tabular images. The fine-tuning employed Low-Rank Adaptation (LoRA) by introducing trainable low-rank matrices into select transformer layers, enabling targeted domain adaptation without requiring full-model retraining. The chosen LoRA configuration involved a rank (r) of 16, a scaling factor (lora_alpha) of 32, and a dropout rate of 0.1 to promote generalization. This configuration aimed to balance effective domain-specific learning with reduced risks of overfitting by limiting the rank and applying regularization through dropout. Figure 12 depicts a snippet of an image sample from the test dataset along with the corresponding question, ground truth answer, and the response generated by the base and fine-tuned Qwen2-VL-2B-Instruct. The figure highlights the impact of fine-tuning on the performance of VLM, with a detailed discussion provided in the subsequent paragraphs.

**Division B**　　　　　　　　　　　　　　　　　　　　　　　　　　　　　　　**9.36.5.10.**

**Table 9.36.5.8.**
**Default Schedule of Service Water Heating Use**
Forming Part of Sentence 9.36.5.8.(6)

| Type of Small Residential Building | Distribution of Hourly Draws on Service Water Heating, L/h | | | | | | | | | | | |
|---|---|---|---|---|---|---|---|---|---|---|---|---|
| Houses without a secondary suite (97 L/day/house) | 12 a.m. | 1 a.m. | 2 a.m. | 3 a.m. | 4 a.m. | 5 a.m. | 6 a.m. | 7 a.m. | 8 a.m. | 9 a.m. | 10 a.m. | 11 a.m. |
| | 0 | 0 | 0 | 0 | 0 | 0 | 0 | 2.2 | 8.6 | 12.9 | 23.7 | 11.9 |
| | 12 p.m. | 1 p.m. | 2 p.m. | 3 p.m. | 4 p.m. | 5 p.m. | 6 p.m. | 7 p.m. | 8 p.m. | 9 p.m. | 10 p.m. | 11 p.m. |
| | 3.2 | 1.1 | 2.2 | 5.4 | 9.7 | 6.5 | 6.5 | 2.2 | 1.1 | 0 | 0 | 0 |
| Each *dwelling unit* in residential *buildings* with two or more *dwelling units* (65 L/day/*dwelling unit*) | 12 a.m. | 1 a.m. | 2 a.m. | 3 a.m. | 4 a.m. | 5 a.m. | 6 a.m. | 7 a.m. | 8 a.m. | 9 a.m. | 10 a.m. | 11 a.m. |
| | 0 | 0 | 0 | 0 | 0 | 0 | 0 | 1.4 | 5.7 | 8.6 | 15.8 | 7.9 |
| | 12 p.m. | 1 p.m. | 2 p.m. | 3 p.m. | 4 p.m. | 5 p.m. | 6 p.m. | 7 p.m. | 8 p.m. | 9 p.m. | 10 p.m. | 11 p.m. |
| | 2.2 | 0.7 | 1.4 | 3.6 | 6.5 | 4.3 | 4.3 | 1.4 | 0.7 | 0 | 0 | 0 |

7) The energy model calculations shall take into account daily hot service water usage for showering
　　a) at 7 a.m. for 15 mins for houses without a *secondary suite*, or
　　b) at 7 a.m. for 10 mins for each *dwelling unit* in residential *buildings* with two or more *dwelling units*.

**Question:** What is the distribution of hourly draws on service water heating at 11 a.m. for houses without a secondary suite?

**Ground Truth Answer:** 11.9 L/h

| Model | Base Generation | Fine-tuned Generation |
|---|---|---|
| LLaMA-3.2-11B-Vision-Instruct | 0 L/h | 11.9 L/h |
| Qwen2-VL-2B-Instruct | 23.7 L/h | 23.7 L/h |
| Qwen2-VL-7B-Instruct | 0 L/h | 11.9 L/h |
| Qwen2.5-VL-3B-Instruct | 11.9 L/h | 11.9 L/h |
| Qwen2.5-VL-7B-Instruct | 23.7 L/h | 11.9 L/h |

**Figure 12.** A sample of a testing dataset with its corresponding question, ground truth answer, and output from both base and fine-tuned Qwen2-VL-2B-Instruct.

After examining qualitative differences between pre-trained and fine-tuned model in Figure 12, Figure 13 presents the quantitative evaluation, showing that the application of LoRA consistently improved model performance across all evaluated VLMs. Notably, relative accuracy gains ranged from 14.29% to 105%, demonstrating LoRA's strong ability to enhance VQA capabilities on tables present in NBCC document. In particular, Qwen2.5-VL-3B-Instruct exhibited the highest relative improvement of 105%, highlighting the effectiveness of lightweight domain adaptation techniques in significantly boosting performance without the need for full-fledged fine-tuning.

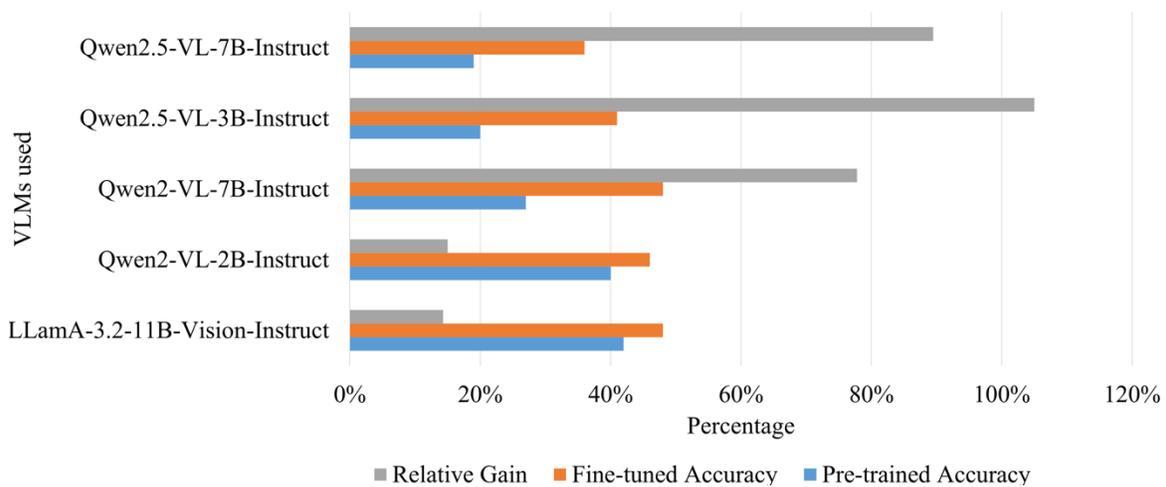

**Figure 13.** Relative gain, accuracy of pre-trained and fine-tuned VLMs

Figure 14 presents the confusion matrices comparing VLM's performance pre- and post-fine-tuning. The matrices illustrate the number of samples correctly or incorrectly predicted by the pre-trained and fine-tuned models. Each subplot corresponds to a different VLM: (a) LLamA-3.2-11B-Vision-Instruct, (b) Qwen2-VL-2B-Instruct, (c) Qwen2-VL-7B-Instruct, (d) Qwen2.5-VL-3B-Instruct, and (e) Qwen2.5-VL-7B-Instruct.

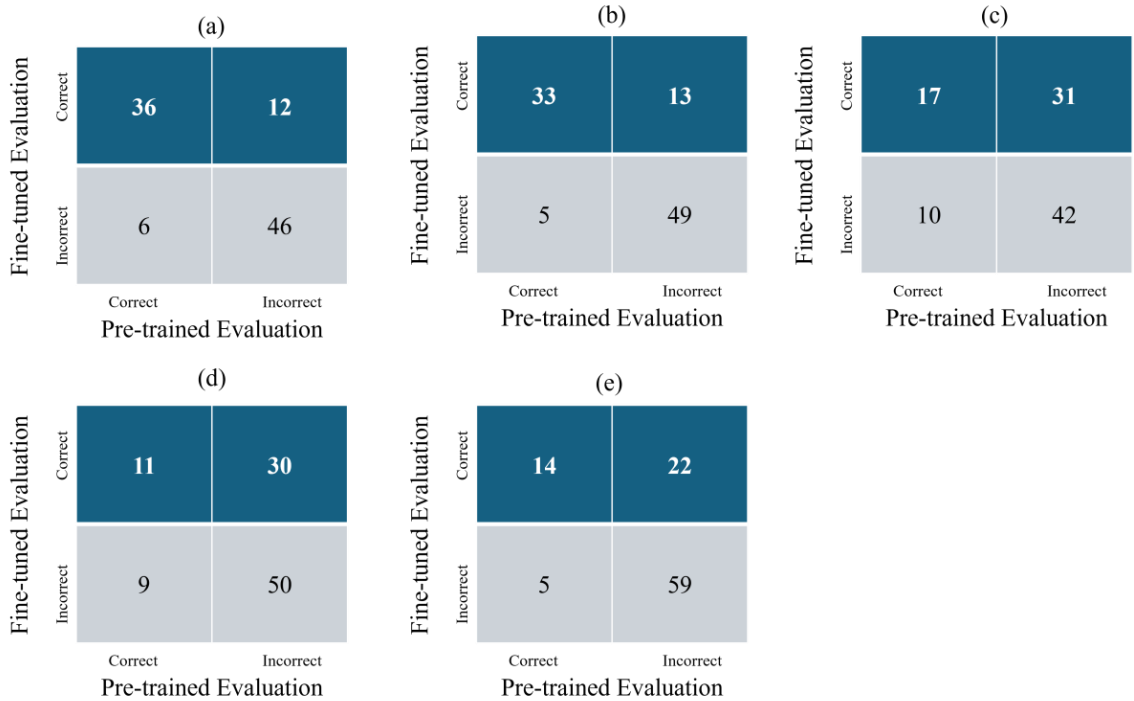

**Figure 14.** Confusion matrix for (a) LLamA-3.2-11B-Vision-Instruct, (b) Qwen2-VL-2B-Instruct, (c) Qwen2-VL-7B-Instruct, (d) Qwen2.5-VL-3B-Instruct, (e) Qwen2.5-VL-7B-Instruct.

The diagonal entries (top-left and bottom-right) indicate agreement between the pre-trained and fine-tuned models. The top-left shows the number of samples where both pre-trained and fine-tuned models predicted a correct response. Bottom right shows the samples where both pre-trained and fine-tuned models were wrong. The top-right indicates the number of samples where the fine-tuned predicted the correct response, and the pre-trained failed to do so. Similarly, the bottom left showed the number of samples where the fine-tuned model failed to answer correctly, but the pre-trained model managed to answer them correctly.

LLamA-3.2-11B-Vision-Instruct (Figure 14a) shows relatively strong performance in both pre-trained and fine-tuned states, achieving 36 correct predictions before and after fine-tuning, and improving further with 12 samples post-fine-tuning. Qwen2-VL-2B-Instruct (Figure 14b) performs similarly, showing consistent gains after fine-tuning with 13 additional correct predictions.

Notably, models with lower pre-trained accuracy demonstrated significant improvement after fine-tuning:

- Qwen2-VL-7B-Instruct (Figure 14c) improved 31 predictions post-fine-tuning. This shift indicates strong adaptability and learning capacity when exposed to supervised instruction data.
- Qwen2.5-VL-3B-Instruct (Figure 14d) dramatically improved on 30 samples after fine-tuning, where pre-trained models failed to answer correctly, showing one of the most significant improvements across all models.
- Qwen2.5-VL-7B-Instruct (Figure 14e) also saw marked enhancement, improving on 22 samples post-fine-tuning, further validating the benefit of LoRA-based fine-tuning for models with initially lower performance.

These results collectively highlight the importance of fine-tuning in boosting model performance, particularly for models that underperform in their pre-trained state. Models with lower pre-trained accuracy tend to receive stronger and more informative gradients during fine-tuning. This results in more significant parameter updates, leading to sharper gains in alignment with downstream tasks.

To further quantify the stability of fine-tuning across models, we computed the ratio between the number of cases where fine-tuning corrected errors by the fine-tuned model and the number of cases where fine-tuning introduced new errors. Table 2 summarizes the correction-to-regression ratio for each model. A higher ratio indicates that fine-tuning improved performance more often than it caused regressions, while a lower ratio suggests greater instability during domain adaptation. Importantly, all models achieved ratios greater than 1, confirming that fine-tuning consistently contributed more to correcting errors than introducing new ones across the entire set of evaluated models. As shown in Table 3, Qwen2.5-VL-7B-Instruct achieved the highest stability ratio of 4.4, followed closely by Qwen2.5-VL-3B-Instruct with a ratio of 3.33, and Qwen2-VL-7B-Instruct with a ratio of 3.1. In contrast, LLaMA-3.2-11B-Vision-Instruct demonstrated the lowest stability ratio of 1.5 among all models evaluated. This indicates that while fine-tuning still resulted in a net improvement for LLaMA-3.2-11B-Vision-Instruct, it also led to relatively higher levels of regression compared to other models. The variability in improvement can largely be attributed to the initial performance of the models in the pre-trained phase using the direct input method. Models that exhibited lower pre-trained accuracy showed a larger gap for improvement and therefore experienced more correction than regression post fine-tuning. Notably, the Qwen2.5 series models demonstrated both substantial accuracy improvements and high fine-tuning stability, as their pre-trained accuracy was very low. On the other hand, models such as LLaMA-3.2-11B-Vision-Instruct and Qwen2 series, which performed well in the pre-training stage, may require more deliberate strategies such as regularization and selective layer tuning to reduce regressions, increase stability, and maximize relative gains post fine-tuning. Additionally, increasing the size of the training dataset could further enhance the effectiveness of fine-tuning and push the boundaries of achievable performance improvements in VLMs.

Table 2. Correction-to-Regression ratio in fine-tuned models.

| Model | Correction-to-Regression Ratio |
| --- | --- |
| LLaMA-3.2-11B-Vision-Instruct | 2.0 |
| Qwen2-VL-2B-Instruct | 2.6 |
| Qwen2-VL-7B-Instruct | 3.1 |
| Qwen2.5-VL-3B-Instruct | 3.33 |
| Qwen2.5-VL-7B-Instruct | 4.4 |

## 5. Conclusion

This study explores the capabilities of VLMs in answering questions based on complex tables from the NBCC and investigates how fine-tuning can enhance their performance. Initially, experiments were conducted to evaluate the effectiveness of different input methods for these models. Among all models tested, LLaMA-3.2-11B-Vision-Instruct performed notably well when using the indirect input method. However, all models, including LLaMA-3.2-11B-Vision-Instruct, demonstrated improved accuracy when the direct input method was employed. Specifically, LLaMA-3.2-11B-Vision-Instruct achieved an accuracy of 42% using the direct input method, compared to 39% with the indirect input method. A similar trend was observed across the other VLMs evaluated. The relatively lower performance with the indirect input method can be attributed to the complexity and abstraction of LaTeX syntax. Additionally, while preparing the dataset for the indirect input method, tables were converted from images to LaTeX using GPT-4.1 and during this conversion process errors may have introduced which could have further contributed to the reduced performance of the VLMs.

VLMs consistently performed better on the direct input method as compared to indirect input method; however, the overall accuracy remained relatively low, particularly for Qwen2.5-3B-Instruct and Qwen2.5-7B-Instruct. To address this, all models were fine-tuned using LoRA. Post fine-tuning, all VLMs demonstrated accuracy improvements, with Qwen2.5-3B showing a relative gain of over 100% compared to its pre-trained version and achieving the highest correction-to-regression ratio of 4.5, indicating strong enhancement in its ability to accurately reason over complex tabular data.

We utilized 400 samples for fine-tuning, which represents one of the limitations of this study. Future work may focus on expanding the training dataset and further fine-tuning the VLMs to achieve greater relative improvements and higher overall accuracy. Although each fine-tuned model demonstrated noticeable performance gains, there remains considerable room for improvement. Hyperparameter optimization could be explored to further enhance model effectiveness. Moreover, since the NBCC contains diverse types of tables serving different purposes, future work could involve classifying these tables into categories and performing detailed analyses for each type. Additionally, a complete end-to-end MRAG framework incorporating the fine-tuned model at its core can be developed and evaluated using a test set curated from various aspects of building codes. Advanced evaluation metrics can then be employed to rigorously assess the performance of different MRAG configurations.

**Acknowledgement**

This research was funded by the Natural Sciences and Engineering Research Council of Canada (NSERC) through the Alliance grant ALLRP 581074-22.

**Data availability**

Dataset used, fine-tuned models and codes will be made available on request.

**Declaration of generative AI and AI-assisted technologies in the writing process**

During the preparation of this work the author used ChatGPT in order to check grammar and paraphrase certain part of document. After grammar check and paraphrasing, the author reviewed and modified the content as needed and take full responsibility for the content of the publication.